%% file: main.tex
\documentclass{article}
\usepackage{xspace}

\PassOptionsToPackage{numbers,compress}{natbib}
\usepackage[final]{neurips}

\input{header}

\newcommand{\nbf}[1]{{\noindent \textbf{#1}}}

\newcommand{\LongWoF}{\texttt{LongWoF-Bench}\xspace}
\newcommand{\gene}{\texttt{Gene}\xspace}
\newcommand{\skill}{\texttt{Skill}\xspace}

\makeatletter
\def\adl@drawiv#1#2#3{%
	\hskip.5\tabcolsep
	\xleaders#3{#2.5\@tempdimb #1{1}#2.5\@tempdimb}%
	#2\z@ plus1fil minus1fil\relax
	\hskip.5\tabcolsep}
\newcommand{\cdashlinelr}[1]{%
	\noalign{\vskip\aboverulesep
		\global\let\@dashdrawstore\adl@draw
		\global\let\adl@draw\adl@drawiv}
	\cdashline{#1}
	\noalign{\global\let\adl@draw\@dashdrawstore
		\vskip\belowrulesep}}
\makeatother

\setlist[itemize]{align=parleft,left=0pt..0.5em}
\setlist[enumerate]{align=parleft,left=0pt..0.5em}

\setlist[itemize]{align=parleft,left=0pt..0.8em}

\newcommand{\evomapheader}{%
  \noindent
  \begin{minipage}[c]{0.62\linewidth}
    \href{https://evomap.ai}{%
      \includegraphics[
        height=14pt,
        keepaspectratio
      ]{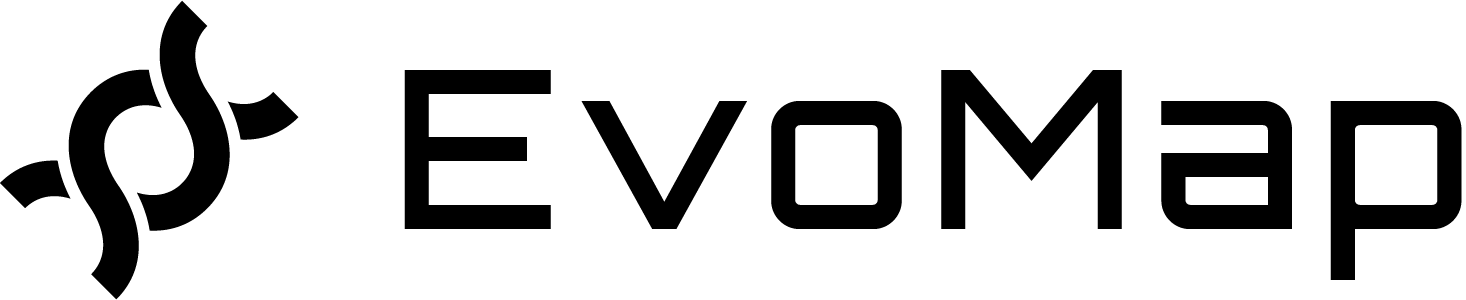}}
  \end{minipage}%
  \hfill
  \vspace{4pt}
}

\def\paperTitle{\includegraphics[height=14pt]{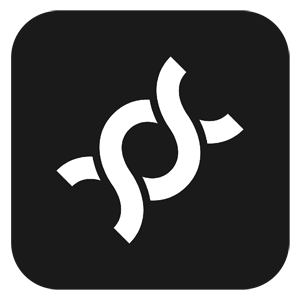} LongWoF-Bench: Evaluating EvoMap Genes for Verifiable Long-Workflow Tasks}

\title{\paperTitle}

\def\authorBlock{
    Xiao Zhang \qquad Qumeng Sun \qquad Jiahao Li \qquad Xiang Liu \qquad  Bruno\\
    Project Leader: \quad \textbf{Haoyang Zhang}\footnotemark[2]
    \\\\\fontsize{10}{10}
    \selectfont{Infinite Evolution Lab, EvoMap}\\
    \includegraphics[height=10pt]{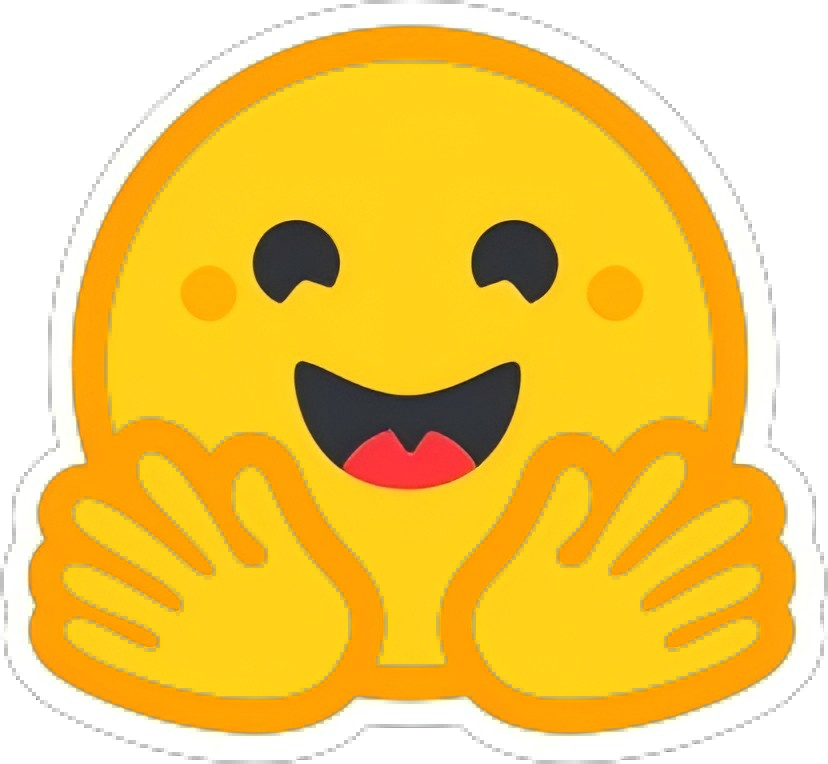} \url{https://huggingface.co/datasets/EvoMapAI/LongWoF-Bench}
}

\author{\authorBlock}

\begin{document}
\evomapheader
\vspace{-0.08in}
\maketitle

{
  \renewcommand{\thefootnote}%
  {\fnsymbol{footnote}}
  \footnotetext[2]{Corresponding Author. {\tt\small 17@evomap.ai}}
}

\begin{abstract}
Large language models are increasingly expected to execute complex workflows whose success depends on maintaining interdependent constraints and producing artifacts that satisfy strict end-to-end verification.
Yet successful execution experience is typically lost after a single run, forcing subsequent models to rediscover strategies and failure modes from scratch.
We study whether such experience can instead be externalized and reused through EvoMap, where verifier-confirmed execution trajectories are consolidated into structured Gene.
To evaluate this setting, we introduce the Long-Workflow Benchmark (\LongWoF), comprising 778 machine-verifiable tasks across code generation, agent-environment synthesis, mathematical reasoning, and rule following.
On the 252 tasks with verifier-confirmed Opus trajectories, evolved EvoMap Gene outperform Skill across all seven evaluated models by 8.7--15.5 percentage points, with the gains extending to consumer models from different model families.
In contrast, reference-distilled Gene do not exhibit the same advantage, indicating that compact representation alone is insufficient and that Gene utility is closely associated with verified experience provenance.
For Claude Opus, Gene reuse also completes 39 more tasks than Skill while reducing solve-time token consumption by 9.9\%.
Together, these results show that verified execution experience can be retained and shared as a reusable external resource, enabling models to improve long-workflow completion without repeatedly paying the full cost of experience discovery.
\end{abstract}

\section{Introduction}

Large language models are increasingly expected to move beyond single-turn problem solving and execute complex end-to-end workflows~\citep{wang2024surveyagents,xi2025riseagents}.
In software implementation, data processing, environment construction, rule-based decision making, and exact computation, success often depends not on whether each individual step appears reasonable, but on whether the entire execution jointly satisfies a set of interdependent constraints~\citep{jimenez2024swebench,boisvert2024workarena,xie2024osworld,sun2025rulebench,hendrycks2021math}.
Later decisions may depend on states, artifacts, or requirements established earlier in the workflow, while a single interface violation, boundary error, or precedence mistake can invalidate an otherwise plausible solution.
\textit{The difficulty of such tasks therefore lies not simply in longer contexts or more execution steps, but in maintaining globally consistent decisions under strict end-to-end requirements.}
We refer to this class of problems as \textit{verifiable long-workflow tasks}, where success is determined by whether the final deliverable satisfies the complete task specification under objective verification.

This setting raises a broader question beyond solving any individual task~\citep{zhao2024expel}: what should happen to the execution experience once a model has successfully completed a difficult workflow?
EvoMap treats such experience as a reusable external asset rather than a transient by-product of a single run~\citep{evomap2026}.
Within this framework, Evolver organizes task execution and, when necessary, feedback-driven refinement until a verifier-confirmed trajectory is obtained~\citep{evomap2026evolver}.
The execution-critical knowledge distilled from this process is consolidated into a structured \gene and stored in EvoMap for subsequent reuse~\citep{wang2026strategygenes,evomap2026}.
\textit{Evolver turns verified execution into reusable experience, while EvoMap makes that experience available beyond the model and run that originally produced it.}

A complementary approach to supporting long workflows is to package reusable procedural knowledge as a \skill, an increasingly common abstraction for equipping agents with task-specific instructions, workflows, and operational conventions~\citep{li2026skillsbench}.
Both \skill and \gene seek to reduce the burden of solving a complex workflow from scratch, but they preserve different forms of knowledge.
A \skill primarily codifies how a task should be carried out, typically emphasizing reusable procedures, interfaces, requirements, and recommended practices.
An evolved \gene instead originates from verifier-confirmed execution and preserves the strategies, corrections, boundary conditions, and failure guards that proved consequential in practice.
\textit{The distinction is therefore not simply one of representation: \skill externalizes procedural knowledge, whereas evolved \gene externalizes verified execution experience.}

For real-world workflows, this distinction raises a practical empirical question: whether reusable procedural knowledge and verifier-grounded execution experience provide different value when models must satisfy the same end-to-end requirements under strict verification.
Existing benchmarks have substantially expanded agent evaluation toward end-to-end tasks in software, web, enterprise, operating-system, and tool-use environments~\citep{jimenez2024swebench,zhou2024webarena,drouin2024workarena,xie2024osworld,yao2024taubench}, while recent studies have begun to explicitly evaluate reusable Skills and procedural memory~\citep{li2026skillsbench,liu2026agenticskills,belikova2026proceduralmemory}.
However, the role of \textit{guidance provenance} remains insufficiently understood, particularly whether experience grounded in verifier-confirmed execution offers advantages over static procedural guidance under otherwise controlled conditions.
To study this question, we introduce the Long-Workflow Benchmark (\LongWoF), comprising 778 machine-verifiable tasks across code generation, agent-environment synthesis, mathematical reasoning, and rule following.
\LongWoF compares No Context, \skill, and EvoMap \gene under shared task specifications, runtimes, and verifiers, enabling a controlled evaluation of how different sources of reusable guidance affect strict end-to-end workflow completion.

Our experiments reveal that the value of reusable guidance depends strongly on how it is acquired.
On the 252 tasks with verifier-confirmed Opus trajectories, evolved EvoMap \gene consistently outperforms \skill across all seven evaluated models, with gains of 8.7--15.5 percentage points in strict pass rate.
These improvements extend beyond the producer model to Sonnet, Gemini, MiniMax, and Qwen families, indicating that verified execution experience can be reused across model families.
In contrast, reference-distilled \gene do not exhibit the same advantage over \skill, suggesting that compact representation alone is insufficient and that guidance provenance plays a critical role.
For Claude Opus, \gene also solves 39 more tasks than \skill while reducing solve-time token consumption by 9.9\%.
Together, these results show that verifier-grounded execution experience can be externalized and reused to improve workflow reliability while reducing repeated exploration.

Our contributions are threefold.
\begin{itemize}
    \item We introduce \LongWoF, a benchmark of 778 machine-verifiable tasks spanning four complementary forms of long-workflow execution under strict end-to-end verification.
    \item We establish a controlled framework for comparing No Context, \skill, and EvoMap \gene, directly testing whether verifier-grounded execution experience provides additional value beyond reusable procedural knowledge.
    \item Across seven models, we show that evolved \gene consistently improve strict task completion over \skill, transfer across model families, and reduce solve-time cost, while further analyses reveal that their effectiveness is strongly associated with experience provenance.
\end{itemize}

\section{\LongWoF}

\subsection{Verifiable Long-Workflow Tasks}

We formulate a verifiable long-workflow task as $\mathcal{T}=(S,E,\mathcal{Y},V)$, where $S$ denotes the public task specification, $E$ the model-accessible environment and resources, $\mathcal{Y}$ the space of admissible deliverables, and $V$ a task-specific machine verifier.
Given $S$ and $E$, a solver performs a sequence of dependent operations and produces a final deliverable $y\in\mathcal{Y}$, which is considered successful only if
\begin{equation}
V(S,E,y)=1.
\end{equation}
The defining property of a long workflow is not its context length, number of actions, or execution time, but the dependency of later decisions on constraints, states, or intermediate artifacts established earlier in the process.
Consequently, locally reasonable decisions can accumulate into an invalid final result when an interface contract, ordering constraint, boundary condition, or required artifact is violated.

Verification in \LongWoF is designed to assess complete task fulfillment rather than intermediate plausibility.
Depending on the task, $V$ is instantiated through executable tests, hidden checks, or normalized exact matching, with success requiring all mandatory conditions to be satisfied.
The information necessary to complete a task remains recoverable from the public specification, accessible environment, and output requirements, while evaluator-only artifacts are used solely to determine whether those requirements have been satisfied.
\textit{A verifiable long-workflow task therefore measures whether a model can preserve interdependent requirements throughout execution and deliver an artifact that survives objective end-to-end verification.}

\subsection{Benchmark Construction and Coverage}

\LongWoF is constructed to place heterogeneous tasks under a common framework of verifiable end-to-end delivery.
Candidate tasks are consolidated into a standardized package containing a model-visible task specification, the required public assets, and a task-specific private evaluator.
Tasks with ambiguous deliverables, insufficiently specified interfaces, unstable execution, or unreliable verification are excluded, such that every retained instance admits an objective and reproducible success criterion.

\begin{figure}[ht]
\centering
\includegraphics[width=0.92\linewidth]{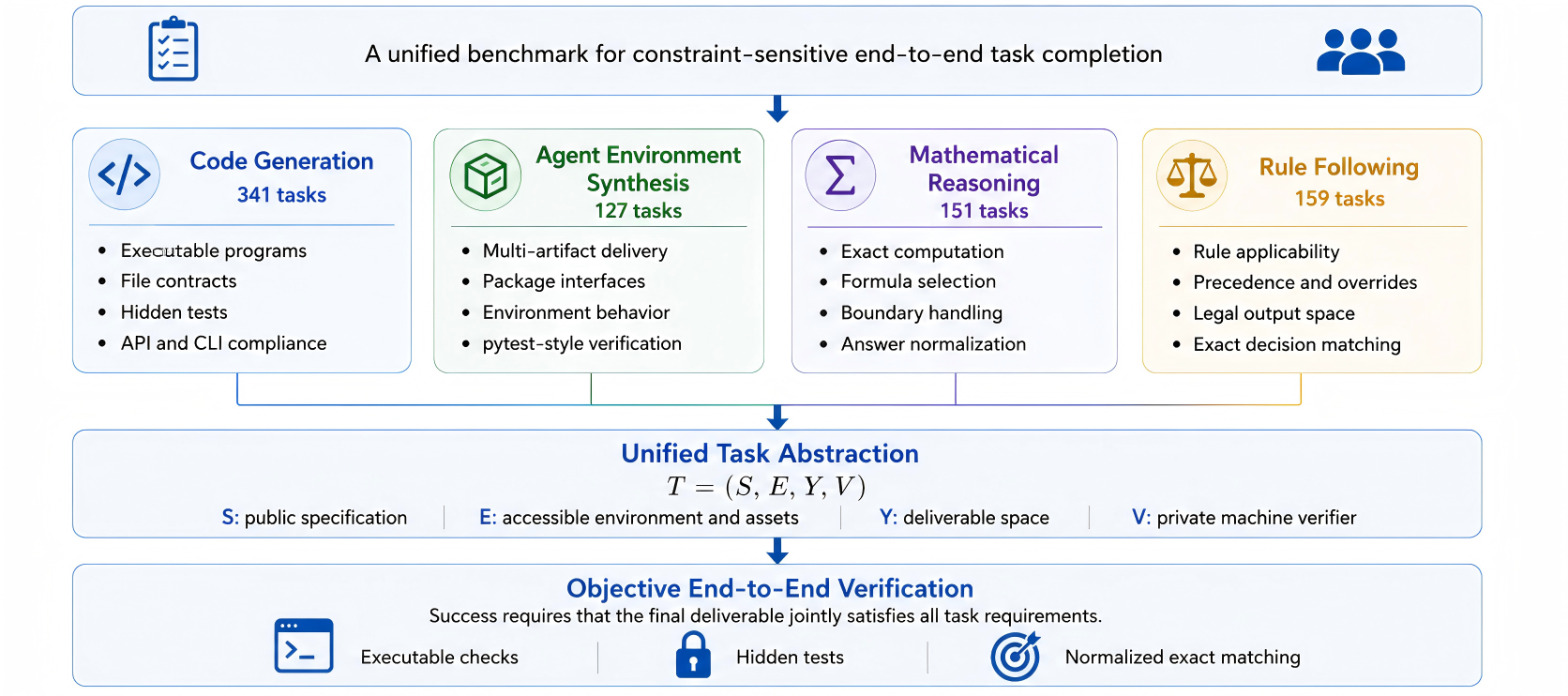}
\caption{Overview of \LongWoF coverage.
The benchmark contains 778 machine-verifiable tasks across four complementary forms of long-workflow execution, unified by a common task abstraction and strict end-to-end verification.}
\label{fig:longwof-benchmark-overview}
\end{figure}

As shown in~\cref{fig:longwof-benchmark-overview}, the current release contains 778 tasks spanning four complementary task families: 341 code-generation tasks, 127 agent-environment-synthesis tasks, 151 mathematical-reasoning tasks, and 159 rule-following tasks.
Code generation emphasizes executable implementations, file contracts, interface compliance, and hidden functional tests.
Agent-environment synthesis further stresses multi-artifact delivery, package interfaces, and environment-level behavior.
Mathematical reasoning requires exact computation and careful treatment of formulas, boundaries, and answer normalization, whereas rule following focuses on applicability, precedence, overrides, and constrained output spaces.
Although these task families differ substantially in their deliverables, each requires models to translate distributed requirements into a final artifact whose correctness can be determined through execution or exact verification.

To make this task structure concrete, \cref{fig:longwof-example} presents a representative agent-environment instance involving the construction of an adaptive cruise-control system.
The public specification provides vehicle parameters, sensor traces, and interface and performance requirements, while successful completion requires coordinated controller implementation, mode switching, parameter tuning, simulation, and multi-artifact delivery.
The submitted system is then independently re-executed by a private evaluator that checks both interface compliance and behavioral constraints.

\begin{figure}[ht]
\centering
\fbox{
\begin{minipage}{0.91\linewidth}
\small
\nbf{Representative Task: Adaptive Cruise-Control System}

\textbf{Public Specification.}
Vehicle parameters, 150-second sensor traces, and interface and performance requirements for cruise, following, and emergency-braking modes.

\textbf{Required Workflow.}
Implement the PID controller and mode-switching logic, tune the controller, execute the simulation, and maintain consistency across generated artifacts.

\textbf{Deliverables.}
Controller and simulator code together with tuning results, simulation outputs, and an analysis report.

\textbf{Private Verification.}
The evaluator imports and executes the submitted implementation, reruns the simulation, checks interfaces and control modes, and validates performance constraints and sensitivity to changed PID parameters.
\end{minipage}
}
\caption{A representative \LongWoF task illustrating how a public specification, dependent execution steps, multiple deliverables, and a private verifier jointly define end-to-end task completion.}
\label{fig:longwof-example}
\end{figure}

Although this example is drawn from agent-environment synthesis, the same benchmark abstraction applies across all four task families: completion requires satisfying publicly specified constraints while correctness is independently determined by a task-specific verifier.
\textit{\LongWoF is therefore unified not by a single application domain, but by a common requirement for constraint-sensitive execution under objective end-to-end verification.}

\subsection{Visibility Boundary and Verification Protocol}

\LongWoF enforces a strict separation between model-visible task information and evaluator-only artifacts.
At evaluation time, the solver receives the public specification, referenced assets, and condition-specific auxiliary context, while hidden tests, gold outputs, reference solutions, verifier logic, and evaluator-only constants remain inaccessible.
Importantly, the private verifier introduces no hidden task requirements: all information necessary for successful completion is recoverable from the public specification, accessible environment, interfaces, and output contract.
Model outputs are materialized as executable artifacts or structured answers and evaluated through program execution, hidden checks, or normalized exact matching.
\textit{A task is counted as successful only when all mandatory verification criteria are satisfied, with the same visibility boundary and verifier held fixed across evaluation conditions.}

\subsection{Workflow Characteristics and Failure Sensitivity}

Although the four task families differ in their deliverables and verification mechanisms, they share a common dependence on constraint-sensitive execution.
Successful completion requires models to preserve interface contracts, ordering relations, boundary conditions, precedence rules, and cross-artifact consistency throughout the workflow.
Many failures therefore arise not from an absence of relevant knowledge, but from locally plausible choices that violate a task-specific requirement when propagated to the final deliverable.
Such errors are particularly consequential under strict verification, where a single incorrect default, omitted artifact, or misapplied boundary can invalidate an otherwise reasonable solution.
\textit{Across \LongWoF, the central challenge is thus to maintain execution-critical constraints until the complete workflow survives objective end-to-end verification.}
This failure sensitivity motivates the study of whether reusable procedural knowledge and verifier-grounded execution experience provide different forms of support, which we examine next through \skill and \gene.

\section{Reusable Experience with EvoMap Genes}

Following the Evolver framework~\citep{evomap2026evolver}, we treat successful task execution as a source of reusable experience rather than an isolated solution.
For each task, a producer model interacts with the verifier under a bounded evolution process, and the resulting verifier-confirmed trajectory is subsequently distilled into a structured \gene.
The resulting asset can then be retained in EvoMap~\citep{evomap2026} and reused by either the original model or a different consumer model.
\textit{This separates the cost of discovering a successful execution strategy from the cost of applying that experience in subsequent runs.}

\subsection{Constructing Genes with Evolver}

We instantiate Gene construction following Evolver's execution--verification--refinement paradigm with GDIv2.
A producer model first attempts the task without access to private verifier information.
If the attempt fails, the next rollout receives the previous solution together with sanitized verifier feedback and revises the solution accordingly.
This process continues within a fixed rollout budget until a passing trajectory is obtained.
Importantly, the object being refined during evolution is the task solution rather than the \gene itself.
Once the verifier accepts a trajectory, we distill its execution-critical information into a structured \gene, preserving successful strategies, prerequisite checks, boundary conditions, and, when applicable, corrections to earlier failures.
For our primary Gene set, Claude Opus serves as the producer model, yielding 252 \gene derived from verifier-confirmed execution trajectories.

\subsection{Reusing and Sharing Genes through EvoMap}

Once produced, a \gene can be reused without replaying the full trajectory or repeating the original verifier-guided refinement process.
For subsequent execution, the consumer receives the public task information together with the corresponding \gene, while the producer trajectory and verifier feedback remain unavailable.
This allows validated experience to support repeated execution by the same model, reducing the need to rediscover task-specific strategies and previously resolved failure modes.

EvoMap further decouples the producer of an experience from its consumer.
A \gene produced from an Claude Opus~\citep{anthropic2026opus48} execution, for example, can be made available to Gemini 3.1 Pro Preview~\citep{google2026gemini31pro}, MiniMax M3~\citep{lai2026minimax}, or other model families as external procedural experience.
\textit{EvoMap therefore enables verified execution experience to persist beyond a single run and to be inherited across models, rather than remaining tied to the agent that originally discovered it.}
The effectiveness and efficiency of these two forms of reuse are evaluated in subsequent sections.

\section{Experimental Setup}

\nbf{Evaluation protocol.}
We evaluate each task under three conditions: No Context, \skill, and EvoMap \gene.
Across conditions, the public task specification, runtime, decoding configuration, and private verifier are held fixed, such that only the auxiliary guidance varies.
Each task--condition pair is evaluated with one recorded trial, and models receive no additional verifier feedback during inference.

\nbf{Models and evaluation sets.}
We evaluate seven models spanning four model families: Claude Opus 4.8 and Claude Sonnet 4.6~\citep{anthropic2026opus48,anthropic2026sonnet46}, Gemini 3.1 Flash-Lite Preview and Gemini 3.1 Pro Preview~\citep{google2026gemini31flashlite,google2026gemini31pro}, MiniMax M3~\citep{lai2026minimax}, Qwen3-Coder-30B-A3B-Instruct~\citep{qwen2025qwen3}, and Qwen3.5-397B-A17B~\citep{qwen2026qwen35}.
Claude Opus 4.8 serves as the primary Gene producer, while all seven models are evaluated as consumers.
The evaluation subsets and their roles are summarized in~\cref{tab:evaluation-sets}.
Our primary comparison focuses on the 252 tasks for which Opus produced verifier-confirmed evolved \gene, while the remaining subsets support analyses of Gene provenance and producer effects.

\begin{table}[ht]
\centering
\caption{Evaluation sets and their roles.
The 252 Opus-evolved and 526 reference-distilled tasks represent different Gene provenance within the full benchmark, while the 180-task common evolved set contains tasks for which both Opus and Gemini produced verifier-confirmed Genes.}
\label{tab:evaluation-sets}
\small
\begin{tabular}{lrl}
\toprule
\textbf{Evaluation set} & \textbf{Tasks} & \textbf{Primary role} \\
\midrule
Full benchmark & 778 & Overall benchmark difficulty \\
Opus-evolved & 252 & Main \gene vs.\ \skill comparison \\
Reference-distilled & 526 & Gene provenance analysis \\
Opus--Gemini common evolved & 180 & Gene producer comparison \\
\bottomrule
\end{tabular}
\vspace{2pt}

\footnotesize
\textit{Note:} One additional task uses a skill-distilled Gene and is included only in full-benchmark reporting.
\end{table}

\nbf{Metrics and statistics.}
Our primary metric is \textit{strict pass rate}, where a task is successful only if all mandatory verifier checks pass.
For efficiency analysis, we additionally report solve-time token consumption, number of model calls, and tokens per passed task.
Paired pass-rate differences are quantified using deterministic task-level bootstrap confidence intervals, with exact McNemar tests used to assess paired significance.
Unless otherwise stated, reported uncertainty reflects variation over tasks rather than repeated hosted-model API runs.

\section{Results}

\subsection{Verifier-Evolved Genes Improve Workflow Completion}

We first evaluate the 252 tasks for which Claude Opus produced a verifier-confirmed trajectory and an evolved \gene.
All three conditions are evaluated on the same tasks with the same private verifier, allowing a direct comparison of how different forms of auxiliary guidance affect strict workflow completion.

\begin{figure}[ht]
\centering
\includegraphics[width=\linewidth]{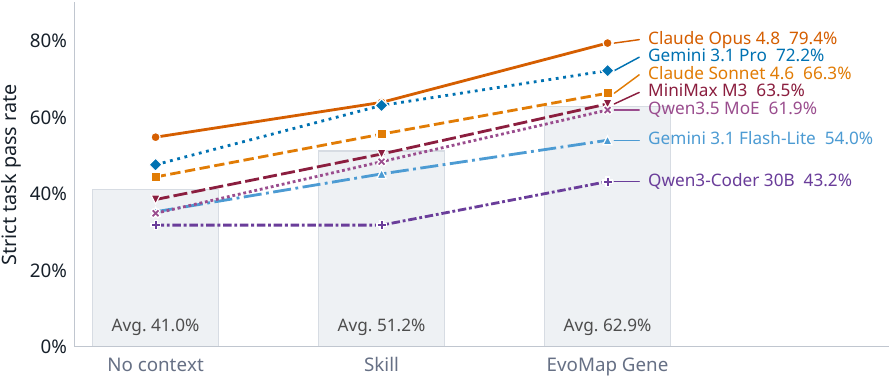}
\caption{Strict task pass rate on the 252 tasks with verifier-evolved Opus Genes.
Each line follows one consumer model across No Context, \skill, and EvoMap \gene, while the background bars show the unweighted mean across the seven models.}
\label{fig:evolved-completion}
\end{figure}

As shown in~\cref{fig:evolved-completion}, reusable guidance improves performance, but evolved EvoMap \gene provides the strongest and most consistent benefit.
Averaged across the seven consumer models, strict pass rate increases from 41.0\% with No Context to 51.2\% with \skill and further to 62.9\% with EvoMap \gene.
More importantly, \gene outperforms \skill for every evaluated model, with gains ranging from 8.7 to 15.5 percentage points.
For Claude Opus 4.8, for example, the pass rate increases from 63.9\% with \skill to 79.4\% with \gene.

The benefit is not restricted to the model that produced the experience.
Although these Genes are derived from Opus trajectories, they also improve Sonnet, Gemini, MiniMax, and Qwen consumers over their corresponding \skill conditions.
\textit{Verifier-confirmed execution experience can therefore remain useful beyond its producer model once it is consolidated into an EvoMap \gene.}

\subsection{Gene Utility Depends on Experience Provenance}

The previous results establish the advantage of verifier-evolved \gene, but leave open whether this benefit follows from the Gene representation itself or from the experience used to construct it.

\begin{table}[ht]
\centering
\small
\caption{Gene utility under verifier-evolved and reference-distilled provenance.
$\Delta$ denotes Gene minus Skill in percentage points.}
\label{tab:provenance-utility}
\resizebox{\linewidth}{!}{%
\begin{tabular}{lrrr@{\hspace{1.2em}}rrr}
\toprule
& \multicolumn{3}{c}{Opus-evolved ($n=252$)} & \multicolumn{3}{c}{Reference-distilled ($n=526$)} \\
\cmidrule(lr){2-4}\cmidrule(lr){5-7}
Model & Skill & Gene & $\Delta$ & Skill & Gene & $\Delta$ \\
\midrule
Anthropic Claude Opus 4.8 & 63.9\% & \textbf{79.4\%} & +15.5 & \textbf{28.2\%} & 19.8\% & -8.4 \\
Anthropic Claude Sonnet 4.6 & 55.6\% & \textbf{66.3\%} & +10.7 & \textbf{22.1\%} & 14.1\% & -8.0 \\
Google Gemini 3.1 Flash-Lite Preview & 45.2\% & \textbf{54.0\%} & +8.7 & \textbf{18.1\%} & 10.9\% & -7.2 \\
Google Gemini 3.1 Pro Preview & 63.1\% & \textbf{72.2\%} & +9.1 & \textbf{26.9\%} & 15.6\% & -11.3 \\
MiniMax M3 & 50.4\% & \textbf{63.5\%} & +13.1 & \textbf{24.0\%} & 16.6\% & -7.4 \\
Qwen/Qwen3-Coder-30B-A3B-Instruct & 31.8\% & \textbf{43.2\%} & +11.5 & \textbf{10.3\%} & 7.0\% & -3.3 \\
Qwen/Qwen3.5-397B-A17B & 48.4\% & \textbf{61.9\%} & +13.5 & \textbf{18.1\%} & 13.5\% & -4.6 \\
\bottomrule
\end{tabular}}
\end{table}

\cref{tab:provenance-utility} reveals a clear provenance-dependent reversal.
On the 252 tasks where Opus obtained a verifier-confirmed trajectory within the evolution budget, \gene outperforms \skill for all seven models by 8.7--15.5 percentage points.
On the complementary 526 tasks where Opus failed to obtain such a trajectory and a fallback \gene was distilled from reference-side teacher signals, \gene instead trails \skill for every model by 3.3--11.3 points.

This contrast suggests that distillation alone cannot substitute for verified experience generation.
Without a verifier-confirmed trajectory, the construction process lacks direct evidence of which plausible execution choices fail and which corrections actually change the verifier outcome.
Effective Gene construction therefore depends not only on how experience is represented, but also on whether the underlying experience has survived end-to-end verification.
Since the two provenance groups contain different tasks, this result should be interpreted as provenance-associated evidence rather than a same-task causal ablation, particularly because the reference-distilled subset consists of tasks that Opus could not solve within the evolution budget.

\begin{figure}[ht]
\centering
\includegraphics[width=\linewidth]{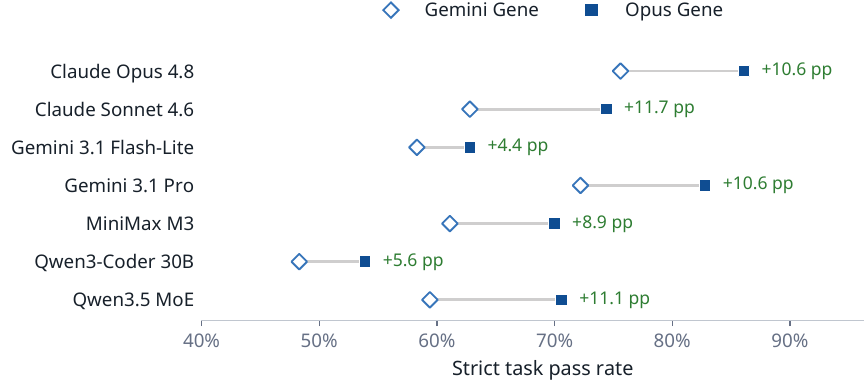}
\caption{Gene producer comparison on the common evolved set ($n=180$).
Each connector compares Gemini-authored and Opus-authored Genes for the same consumer model, with labels reporting the Opus-minus-Gemini difference in strict pass rate.}
\label{fig:gene-author}
\end{figure}

Gene provenance also varies with the model that produces the underlying experience.
To control for task selection, we restrict this analysis to the 180 tasks for which both Opus and Gemini produced verifier-confirmed trajectories.
As shown in~\cref{fig:gene-author}, Opus-authored \gene outperform Gemini-authored \gene for every consumer model, with gains ranging from 4.4 to 11.7 percentage points.
Because the task set is fixed, this result provides stronger evidence that the quality of the producing process affects downstream Gene utility.
It does not, however, isolate whether the difference arises from exploration quality, the experience selected during distillation, or how that experience is expressed in the final \gene.
\textit{Verified provenance matters not only in whether a Gene is grounded in successful execution, but also in the quality of the execution experience from which it is produced.}

\subsection{Gene Reuse Reduces Repeated Exploration}

We next examine whether previously discovered execution experience can reduce the cost of solving the same workflows again.
Under the same one-shot protocol on the 252 evolved tasks, \skill passes 161 tasks using 803,099 solve-time tokens, whereas EvoMap \gene passes 200 tasks using 723,480 tokens.
Thus, \gene completes 39 additional tasks while consuming 79,619 fewer tokens, corresponding to a 9.9\% reduction in total solve-time cost.
\textit{Gene reuse therefore improves both task completion and inference efficiency relative to static procedural guidance.}

\begin{figure}[ht]
\centering
\includegraphics[width=0.9\linewidth]{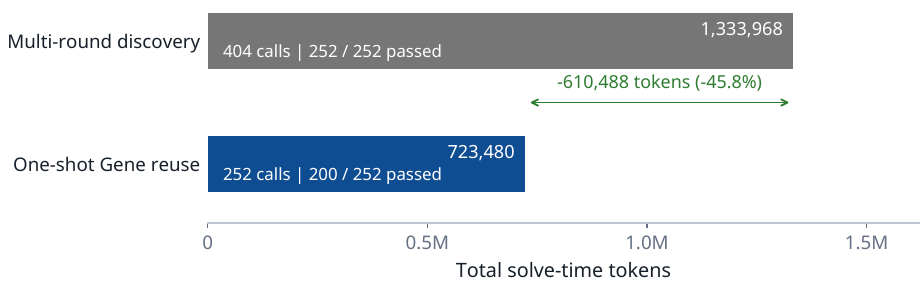}
\caption{Solve-time token cost of multi-round experience discovery and one-shot Gene reuse on the same 252 tasks.
Discovery can consume verifier feedback across multiple calls, whereas reuse performs a single Gene-guided attempt per task.}
\label{fig:discovery-reuse-cost}
\end{figure}

\cref{fig:discovery-reuse-cost} further compares one-shot reuse with the original process used to discover the verified experience.
Multi-round discovery requires 404 model calls and 1,333,968 solve-time tokens to obtain verifier-confirmed trajectories for all 252 tasks, whereas one-shot Gene reuse requires 252 calls and 723,480 tokens, reducing solve-time tokens by 45.8%.
Since discovery can repeatedly consume verifier feedback and ultimately solves all 252 tasks, while one-shot reuse passes 200, this comparison measures the amortized cost of reusing previously discovered experience rather than an equal-success replacement.
\textit{EvoMap does not eliminate the cost of discovering successful experience; it allows that cost to be amortized across subsequent executions.}

\begin{table}[ht]
\centering
\small
\caption{Discovery and one-shot Opus-Gene reuse cost by workflow family.
Change denotes the reduction in average solve-time tokens per task.}
\label{tab:token-by-family}
\resizebox{\linewidth}{!}{%
\begin{tabular}{lrrrrrrrr}
\toprule
Family & Tasks & Disc. calls & Disc. total & Disc. avg. & Gene passed & Gene total & Gene avg. & Change \\
\midrule
Agent environment & 64 & 90 & 310,624 & 4853.5 & 53 & 165,110 & 2579.8 & -46.8\% \\
Code generation & 62 & 100 & 562,667 & 9075.3 & 43 & 248,955 & 4015.4 & -55.8\% \\
Math reasoning & 30 & 42 & 264,794 & 8826.5 & 25 & 176,783 & 5892.8 & -33.2\% \\
Rule following & 96 & 172 & 195,883 & 2040.4 & 79 & 132,632 & 1381.6 & -32.3\% \\
\midrule
Total & 252 & 404 & 1,333,968 & 5293.5 & 200 & 723,480 & 2871.0 & -45.8\% \\
\bottomrule
\end{tabular}%
}
\end{table}

As shown in~\cref{tab:token-by-family}, reuse reduces average solve-time tokens across all four workflow families, with the largest savings in code generation (55.8\%) and agent-environment synthesis (46.8\%).
These tasks frequently require repeated construction or revision of executable artifacts, making the cost of rediscovering an execution strategy particularly high.
The reported accounting includes all one-shot Gene attempts, including failures, but excludes the one-time cost of Gene distillation and auditing.

\subsection{Gains Vary Across Workflow Types}

\begin{figure}[ht]
\centering
\includegraphics[width=0.98\linewidth]{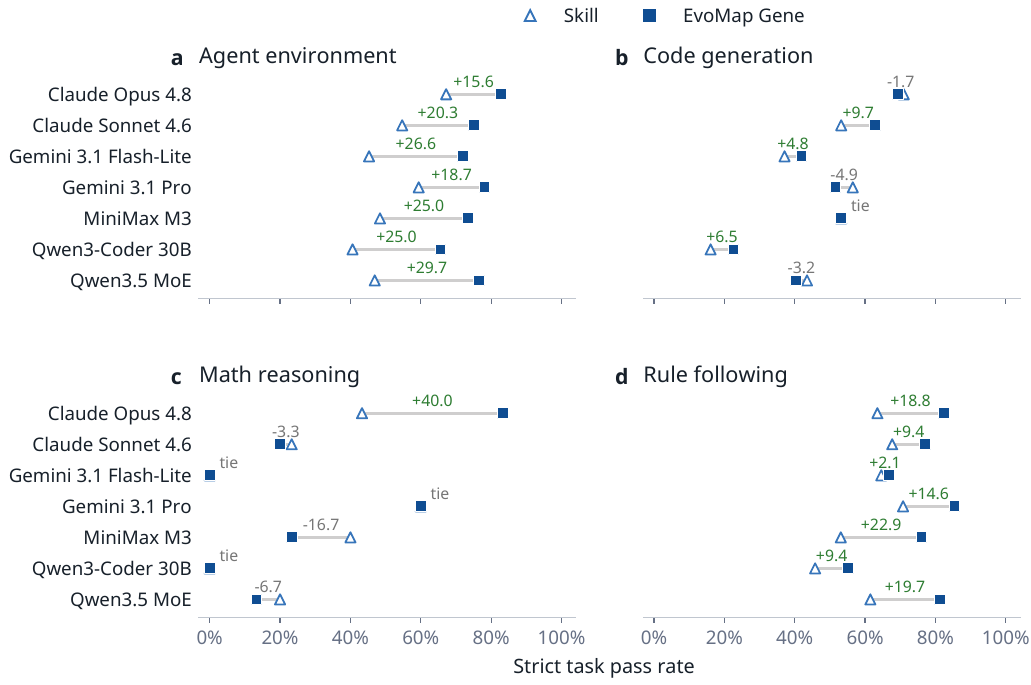}
\caption{Skill and Gene pass rates by workflow family on the 252-task evolved subset.
Signed labels report Gene-minus-Skill differences in percentage points, with negative values and ties retained.}
\label{fig:workflow-type-gains}
\end{figure}

The benefit of evolved \gene varies across workflow types.
As shown in~\cref{fig:workflow-type-gains}, the most consistent gains occur in agent-environment synthesis and rule following, where \gene improves all seven consumer models by 15.6--29.7 and 2.1--22.9 percentage points, respectively.
These tasks frequently depend on fragile operational conventions, including interface and multi-artifact consistency, precedence, overrides, and boundary conditions.
\textit{Verified execution experience appears particularly useful when success depends on preserving task-specific operational constraints throughout the workflow.}

Code generation is more model-dependent, although the observed regressions are relatively small: the three negative differences range from only 1.7 to 4.9 points, while the remaining models show gains or a tie.
This variation suggests that some coding tasks benefit from the broader interface and implementation coverage of \skill, whereas others depend on a small number of execution conventions that are more effectively preserved by \gene.
Mathematical reasoning shows a different pattern, with a substantial gain for Opus but ties or regressions for several other models.
Unlike interface, precedence, or workflow errors, many mathematical failures ultimately depend on whether the consumer model can carry out exact multi-step reasoning and computation after the relevant procedure has been identified.
Gene can transfer formulas, boundary conventions, and solution strategies, but it cannot fully compensate when the dominant bottleneck lies in the model's underlying reasoning capability.
These results suggest that Gene utility may also be influenced by the interaction between the transferred experience, the target model, and the structure of the workflow.

\subsection{Case Studies: Transferring Execution-Critical Experience}

Aggregate results show when evolved \gene improve performance, while representative cases illustrate what execution-critical information is actually transferred.
We focus on two tasks that expose different failure modes, with complete trajectories and additional cases provided in the Appendix.

\nbf{T0161: Drill-Hole Assay Compositing.}
The task requires overlapping assay intervals to be resolved before compositing them into fixed-length bins.
Opus initially adopts a plausible ``later interval wins'' convention, but this strategy repeatedly fails the verifier.
The successful trajectory instead segments intervals at every breakpoint, averages values from all assays covering each segment, and then applies length-weighted aggregation.
Without the \gene, Gemini Pro follows a similarly incorrect truncation strategy and produces an invalid mineralization score; with the evolved \gene, it applies the verified overlap semantics and passes in one attempt.
\textit{The transferred experience is therefore not a generic instruction to handle overlaps, but the specific semantic convention that proved necessary under verification.}

\nbf{T0659: Emergency Dispatch.}
This task combines a strict distance threshold with precedence among injury, armed-without-injury, and distance-only rules.
Both an intermediate Opus rollout and unguided Gemini Pro over-escalate the case by treating weapon presence as dominant and mishandling the equality boundary.
The verifier-confirmed trajectory resolves both issues by treating equality as non-triggering and applying the correct precedence order.
With this experience encoded in the \gene, Gemini Pro avoids the same failure and returns the verified dispatch decision.
\textit{Together, these cases show how evolved Genes can transfer the precise corrections that distinguish a plausible execution path from one that survives end-to-end verification.}

\section{Conclusion}

We introduced \LongWoF, a benchmark of 778 machine-verifiable tasks for evaluating strict end-to-end completion under interdependent workflow constraints.
Using this benchmark, we studied whether execution experience consolidated as EvoMap \gene provides value beyond reusable procedural knowledge represented by \skill.
On the tasks with verifier-confirmed Opus trajectories, evolved \gene consistently outperform \skill across all seven evaluated consumer models, demonstrating that verified execution experience can remain useful beyond the model that originally produced it.
Our provenance analysis further shows that this advantage does not arise from the Gene representation alone, as reference-distilled \gene do not exhibit the same behavior.
Gene reuse also reduces solve-time cost relative to both static \skill guidance and the original multi-round discovery process, indicating that successful exploration can be amortized across subsequent executions.
Taken together, these results suggest that verified agent experience can be externalized, retained, and shared through EvoMap, turning successful task execution from a transient outcome into a reusable resource for future models and workflows.

\bibliographystyle{unsrtnat}
\bibliography{references}

\clearpage
\appendix
\section*{Appendix}

\section{Detailed Experimental Results}

\subsection{Main Evolved-Gene Results}

This section provides the complete numerical results underlying the main evolved-Gene comparison in~\cref{fig:evolved-completion}.
As shown in~\cref{tab:app-evolved-results}, \gene achieves the highest strict pass rate for all seven consumer models on the same 252 Opus-evolved tasks.
The corresponding paired effects are reported in~\cref{tab:app-evolved-effects}, where all bootstrap confidence intervals exclude zero and all improvements are significant under exact McNemar tests.

\begin{table}[ht]
\centering
\small
\caption{Strict pass rate on the 252 tasks with an Opus-evolved \gene.}
\label{tab:app-evolved-results}
\begin{tabular}{lrrr}
\toprule
\textbf{Model} & \textbf{No Context} & \textbf{\skill} & \textbf{Opus \gene} \\
\midrule
Anthropic Claude Opus 4.8 & 54.8\% & 63.9\% & \textbf{79.4\%} \\
Anthropic Claude Sonnet 4.6 & 44.4\% & 55.6\% & \textbf{66.3\%} \\
Google Gemini 3.1 Flash-Lite Preview & 35.3\% & 45.2\% & \textbf{54.0\%} \\
Google Gemini 3.1 Pro Preview & 47.6\% & 63.1\% & \textbf{72.2\%} \\
MiniMax M3 & 38.5\% & 50.4\% & \textbf{63.5\%} \\
Qwen/Qwen3-Coder-30B-A3B-Instruct & 31.8\% & 31.8\% & \textbf{43.2\%} \\
Qwen/Qwen3.5-397B-A17B & 34.9\% & 48.4\% & \textbf{61.9\%} \\
\bottomrule
\end{tabular}
\end{table}

\begin{table}[ht]
\centering
\small
\caption{Paired Opus-\gene minus \skill effects on the 252-task evolved subset.
Confidence intervals use a deterministic paired task bootstrap, and significance levels are based on exact McNemar tests.}
\label{tab:app-evolved-effects}
\resizebox{\linewidth}{!}{%
\begin{tabular}{lrrr}
\toprule
\textbf{Model} & \textbf{$\Delta$ (pp)} & \textbf{Paired Bootstrap 95\% CI} & \textbf{Significance} \\
\midrule
Anthropic Claude Opus 4.8 & 15.48 & [8.33, 22.22] & $p<0.001$ \\
Anthropic Claude Sonnet 4.6 & 10.71 & [3.57, 17.46] & $p<0.01$ \\
Google Gemini 3.1 Flash-Lite Preview & 8.73 & [1.98, 15.87] & $p<0.05$ \\
Google Gemini 3.1 Pro Preview & 9.13 & [2.38, 15.49] & $p<0.01$ \\
MiniMax M3 & 13.10 & [5.94, 20.24] & $p<0.001$ \\
Qwen/Qwen3-Coder-30B-A3B-Instruct & 11.51 & [4.37, 18.65] & $p<0.01$ \\
Qwen/Qwen3.5-397B-A17B & 13.49 & [6.35, 20.24] & $p<0.001$ \\
\bottomrule
\end{tabular}%
}
\end{table}

\subsection{Gene Provenance and Producer Effects}

This section reports the complete results underlying the provenance and Gene-producer analyses in the main text.
As shown in~\cref{tab:app-reference-results}, reference-distilled \gene trail \skill across all seven consumer models on the complementary 526-task subset.
\cref{tab:app-common-results} further reports results on the 180 tasks for which both Opus and Gemini produced verifier-confirmed trajectories, providing a task-matched comparison of Gene producers.
Across this common set, Opus-authored \gene outperform Gemini-authored \gene for every consumer model.

\begin{table}[ht]
\centering
\small
\caption{Strict pass rate on the 526 tasks with a reference-distilled Opus \gene.}
\label{tab:app-reference-results}
\begin{tabular}{lrrr}
\toprule
\textbf{Model} & \textbf{No Context} & \textbf{\skill} & \textbf{Opus \gene} \\
\midrule
Anthropic Claude Opus 4.8 & 3.6\% & \textbf{28.2\%} & 19.8\% \\
Anthropic Claude Sonnet 4.6 & 3.2\% & \textbf{22.1\%} & 14.1\% \\
Google Gemini 3.1 Flash-Lite Preview & 2.1\% & \textbf{18.1\%} & 10.9\% \\
Google Gemini 3.1 Pro Preview & 4.6\% & \textbf{26.9\%} & 15.6\% \\
MiniMax M3 & 4.2\% & \textbf{24.0\%} & 16.6\% \\
Qwen/Qwen3-Coder-30B-A3B-Instruct & 2.5\% & \textbf{10.3\%} & 7.0\% \\
Qwen/Qwen3.5-397B-A17B & 2.5\% & \textbf{18.1\%} & 13.5\% \\
\bottomrule
\end{tabular}%
\end{table}

\begin{table}[ht]
\centering
\small
\caption{Complete results on the 180 tasks for which both Opus and Gemini produced verifier-confirmed \gene.
$\Delta$ denotes Opus \gene minus Gemini \gene in percentage points.}
\label{tab:app-common-results}
\resizebox{\linewidth}{!}{%
\begin{tabular}{lrrrrr}
\toprule
\textbf{Model} & \textbf{No Context} & \textbf{\skill} & \textbf{Gemini \gene} & \textbf{Opus \gene} & \textbf{$\Delta$} \\
\midrule
Anthropic Claude Opus 4.8 & 61.7\% & 68.9\% & 75.6\% & \textbf{86.1\%} & +10.6 \\
Anthropic Claude Sonnet 4.6 & 53.3\% & 61.1\% & 62.8\% & \textbf{74.4\%} & +11.7 \\
Google Gemini 3.1 Flash-Lite Preview & 45.6\% & 50.6\% & 58.3\% & \textbf{62.8\%} & +4.4 \\
Google Gemini 3.1 Pro Preview & 63.3\% & 71.1\% & 72.2\% & \textbf{82.8\%} & +10.6 \\
MiniMax M3 & 46.7\% & 57.8\% & 61.1\% & \textbf{70.0\%} & +8.9 \\
Qwen/Qwen3-Coder-30B-A3B-Instruct & 42.2\% & 37.2\% & 48.3\% & \textbf{53.9\%} & +5.6 \\
Qwen/Qwen3.5-397B-A17B & 41.7\% & 54.4\% & 59.4\% & \textbf{70.6\%} & +11.1 \\
\bottomrule
\end{tabular}%
}
\end{table}

\subsection{Experience Discovery and Reuse Cost}

This section provides the detailed cost accounting underlying the experience-discovery and reuse analysis in~\cref{fig:discovery-reuse-cost,tab:token-by-family}.
As shown in~\cref{tab:app-evolution-depth}, 140 of the 252 verifier-confirmed trajectories are discovered on the first attempt, while 112 tasks require at least one feedback-driven refinement.
These refinements introduce 152 additional model calls beyond a one-call-per-task baseline, illustrating the trial-and-error cost that a reusable \gene can subsequently avoid.

\begin{table}[ht]
\centering
\small
\caption{Exploration rounds required to discover the 252 verifier-confirmed Opus trajectories.
Additional calls are measured relative to one attempt per task.}
\label{tab:app-evolution-depth}
\begin{tabular}{lrrrr}
\toprule
\textbf{Passing Round} & \textbf{Thinking} & \textbf{Tasks} & \textbf{Share} & \textbf{Additional Calls} \\
\midrule
1 & off & 140 & 55.6\% & 0 \\
2 & low & 72 & 28.6\% & 72 \\
3 & high & 40 & 15.9\% & 80 \\
\midrule
Total & off/low/high & 252 & 100.0\% & 152 \\
\bottomrule
\end{tabular}
\end{table}

\cref{tab:app-token-efficiency} reports the corresponding end-to-end solve-time cost on the same 252 tasks.
Multi-round discovery reaches all 252 verified solutions using 404 calls and 1,333,968 tokens, whereas one-shot Opus-\gene reuse passes 200 tasks using 252 calls and 723,480 tokens.
Under the same one-shot protocol, \gene also improves over \skill in both effectiveness and efficiency, passing 39 additional tasks while using 9.9\% fewer tokens.
The accounting includes all solve attempts, including failures, but excludes the one-time cost of Gene distillation and auditing.

\begin{table}[ht]
\centering
\small
\caption{Solve-time cost on the same 252 evolved tasks.
Multi-round discovery can use verifier feedback, whereas the remaining conditions make one call per task.}
\label{tab:app-token-efficiency}
\begin{tabular}{lrrrrrr}
\toprule
\textbf{Mode} & \textbf{Calls} & \textbf{Passed} & \textbf{Pass Rate} & \textbf{Total Tokens} & \textbf{Avg./Task} & \textbf{Tokens/Pass} \\
\midrule
Multi-round discovery & 404 & 252 & 100.0\% & 1,333,968 & 5293.5 & 5293.5 \\
No Context, single call & 252 & 138 & 54.8\% & 651,388 & 2584.9 & 4720.2 \\
\skill, single call & 252 & 161 & 63.9\% & 803,099 & 3186.9 & 4988.2 \\
Opus \gene, single call & 252 & 200 & 79.4\% & 723,480 & 2871.0 & 3617.4 \\
\bottomrule
\end{tabular}%
\end{table}

\subsection{Results by Workflow Type}

This section provides the complete workflow-family results underlying~\cref{fig:workflow-type-gains}.
\cref{tab:app-evolved-tracks} reports No Context, \skill, and Opus-\gene strict pass rates for all seven consumer models on the 252 evolved tasks.
The full breakdown preserves the heterogeneous patterns discussed in the main text, including consistently positive Gene effects for agent-environment synthesis and rule following, as well as the more model-dependent results in code generation and mathematical reasoning.

\begin{table}[p]
\centering
\scriptsize
\caption{Complete workflow-family breakdown on the 252 tasks with an Opus-evolved \gene.}
\label{tab:app-evolved-tracks}
\resizebox{\linewidth}{!}{%
\begin{tabular}{llrrr}
\toprule
\textbf{Family} & \textbf{Model} & \textbf{No Context} & \textbf{\skill} & \textbf{Opus \gene} \\
\midrule
Agent environment
& Anthropic Claude Opus 4.8 & 75.0\% & 67.2\% & \textbf{82.8\%} \\
& Anthropic Claude Sonnet 4.6 & 67.2\% & 54.7\% & \textbf{75.0\%} \\
& Google Gemini 3.1 Flash-Lite Preview & 56.2\% & 45.3\% & \textbf{71.9\%} \\
& Google Gemini 3.1 Pro Preview & 65.6\% & 59.4\% & \textbf{78.1\%} \\
& MiniMax M3 & 51.6\% & 48.4\% & \textbf{73.4\%} \\
& Qwen/Qwen3-Coder-30B-A3B-Instruct & 54.7\% & 40.6\% & \textbf{65.6\%} \\
& Qwen/Qwen3.5-397B-A17B & 54.7\% & 46.9\% & \textbf{76.6\%} \\
\midrule
Code generation
& Anthropic Claude Opus 4.8 & 56.5\% & \textbf{71.0\%} & 69.3\% \\
& Anthropic Claude Sonnet 4.6 & 38.7\% & 53.2\% & \textbf{62.9\%} \\
& Google Gemini 3.1 Flash-Lite Preview & 22.6\% & 37.1\% & \textbf{41.9\%} \\
& Google Gemini 3.1 Pro Preview & 30.6\% & \textbf{56.5\%} & 51.6\% \\
& MiniMax M3 & 29.0\% & \textbf{53.2\%} & \textbf{53.2\%} \\
& Qwen/Qwen3-Coder-30B-A3B-Instruct & 11.3\% & 16.1\% & \textbf{22.6\%} \\
& Qwen/Qwen3.5-397B-A17B & 27.4\% & \textbf{43.5\%} & 40.3\% \\
\midrule
Mathematical reasoning
& Anthropic Claude Opus 4.8 & 43.3\% & 43.3\% & \textbf{83.3\%} \\
& Anthropic Claude Sonnet 4.6 & 16.7\% & \textbf{23.3\%} & 20.0\% \\
& Google Gemini 3.1 Flash-Lite Preview & 0.0\% & 0.0\% & 0.0\% \\
& Google Gemini 3.1 Pro Preview & \textbf{63.3\%} & 60.0\% & 60.0\% \\
& MiniMax M3 & 33.3\% & \textbf{40.0\%} & 23.3\% \\
& Qwen/Qwen3-Coder-30B-A3B-Instruct & 0.0\% & 0.0\% & 0.0\% \\
& Qwen/Qwen3.5-397B-A17B & 0.0\% & \textbf{20.0\%} & 13.3\% \\
\midrule
Rule following
& Anthropic Claude Opus 4.8 & 43.8\% & 63.5\% & \textbf{82.3\%} \\
& Anthropic Claude Sonnet 4.6 & 41.7\% & 67.7\% & \textbf{77.1\%} \\
& Google Gemini 3.1 Flash-Lite Preview & 40.6\% & 64.6\% & \textbf{66.7\%} \\
& Google Gemini 3.1 Pro Preview & 41.7\% & 70.8\% & \textbf{85.4\%} \\
& MiniMax M3 & 37.5\% & 53.1\% & \textbf{76.0\%} \\
& Qwen/Qwen3-Coder-30B-A3B-Instruct & 39.6\% & 45.8\% & \textbf{55.2\%} \\
& Qwen/Qwen3.5-397B-A17B & 37.5\% & 61.5\% & \textbf{81.2\%} \\
\bottomrule
\end{tabular}%
}
\end{table}

\end{document}

%% file: header.tex
\usepackage{latexsym}
\usepackage{tcolorbox}

\usepackage[T1]{fontenc}
\usepackage[utf8]{inputenc}
\usepackage{microtype}

\usepackage{inconsolata}
\usepackage{wrapfig,lipsum,booktabs}
\usepackage{pifont}
\usepackage{makecell} %multiline cell
\usepackage{tabularx}
\usepackage[normalem]{ulem}
\useunder{\uline}{\ul}{}
\usepackage[export]{adjustbox}

\usepackage{hyperref}       % hyperlinks
\usepackage{url}            % simple URL typesetting
\usepackage{booktabs}       % professional-quality tables
\usepackage{amsfonts}       % blackboard math symbols
\usepackage{nicefrac}       % compact symbols for 1/2, etc.

\usepackage{graphicx}
\usepackage{subfigure}
\usepackage{float}
\usepackage{multirow}

\usepackage{listings}

\usepackage{amsmath}
\usepackage{amssymb}
\usepackage{mathtools}
\usepackage{amsthm}
\usepackage{balance}
\usepackage{flushend}

\usepackage{lipsum}
\usepackage{graphicx}
\usepackage{subcaption}
\usepackage{tikz}
\usetikzlibrary{shapes, arrows, positioning}

\usepackage{fancyhdr}

\hypersetup{
	colorlinks,
	citecolor=gray,
	linkcolor=red,
	urlcolor=blue}

\usepackage[capitalize,noabbrev]{cleveref}
\usepackage{tablefootnote}

\usepackage{color, colortbl,xcolor}
\usepackage{enumitem}
\usepackage{booktabs,arydshln}